\documentclass{article}

 \usepackage[eandd, final]{neurips_2026}

\usepackage[utf8]{inputenc} 
\usepackage[T1]{fontenc}    
\usepackage{hyperref}       
\usepackage{amsfonts}       
\usepackage{nicefrac}       
\usepackage{microtype}      
\usepackage{xcolor}         

\usepackage{makecell}
\usepackage{siunitx}
\usepackage{animate}
\usepackage{graphicx}	
\usepackage{amsmath}	
\usepackage{amssymb}	
\usepackage{amsthm}
\usepackage{booktabs}
\usepackage{times}
\usepackage{epsfig}
\usepackage{caption}
\usepackage{float}
\usepackage{placeins}
\usepackage{color, colortbl}
\usepackage{stfloats}
\usepackage{enumitem}
\usepackage{tabularx}
\usepackage{xstring}
\usepackage{multirow}
\usepackage{xspace}
\usepackage{url}
\usepackage{subcaption}
\usepackage[hang,flushmargin]{footmisc}
\usepackage{adjustbox}
\usepackage{wrapfig}
\usepackage{algorithm,algorithmicx,algpseudocode}
\usepackage{listings}
\usepackage{bm}

\usepackage{pifont}
\usepackage{thmtools,thm-restate}

\definecolor{tabfirst}{rgb}{1, 0.7, 0.7}
\definecolor{tabsecond}{rgb}{1, 0.85, 0.7}
\definecolor{tabthird}{rgb}{1, 1, 0.7}

\newcommand{\Oc}{{\mathcal{O}}}

\newcommand{\Vc}{{\mathcal{V}}}

\def\[#1\]{\begin{align}#1\end{align}}

\definecolor{GoogleBlue}{RGB}{66, 133, 244}
\definecolor{GoogleRed}{RGB}{234, 67, 53}
\definecolor{GoogleYellow}{RGB}{251, 188, 4}
\definecolor{GoogleGreen}{RGB}{52, 168, 83}

\usepackage[most]{tcolorbox}

\newtcolorbox[auto counter]{finding}[1][]{%
  enhanced,
  breakable,
  colback=white,
  colframe=black!25,
  boxrule=0.6pt,
  arc=2pt,
  left=6pt,right=6pt,top=5pt,bottom=5pt,   
  before skip=4pt,
  after skip=4pt,
  borderline west={2pt}{0pt}{black!25},
  before upper={\textbf{Finding~\thetcbcounter.}~},
  #1
}

\title{MVVBench: Benchmarking 4D Reasoning in Vision-Language Models}

\author{%
    Hyungjin Chung$^{1,2*}$,
    Byeongjun Park$^{1*}$,
    Joonseok Lee$^{1}$,
    Hojun Kim$^{1}$,
    Jaeho Choi$^{1}$,\\
    \textbf{Byung-Hoon Kim}$^{1,3}$\\
    $^{1}$EverEx \quad $^{2}$Korea University \quad $^{3}$Yonsei University \\
}

\begin{document}

\maketitle

\let\thefootnote\relax\footnotetext{%
  $^*$: Equal contribution.
}

\begin{abstract}
Multi-view video understanding requires integrating spatial and temporal evidence across multiple, often non-overlapping camera streams—tracking entities as they transition between viewpoints, aligning events across time, and reasoning about latent 4D continuity rather than any single visible frame. 
We introduce MVVBench, a benchmark for multi-view video reasoning built from real-world multi-camera datasets. Questions are curated to be monocular-ambiguous along both the view and the temporal axis: each question is unanswerable from any single view in the designated input set, and the majority are further unanswerable from any single moment. Each question becomes uniquely solvable only by jointly reasoning across views and across time.
MVVBench spans diverse dynamic scenes and probes six capabilities: implicit/explicit attribute identification, implicit/explicit relative distance, relative camera pose, and compositional counting, with human-authored QA and rigorous verification.
Beyond benchmarking, we provide an extensive analysis of when and why current vision-language models succeed or fail, characterizing errors due to temporal mis-localization, cross-view identity breaks, and brittle multi-hop reasoning. We then study inference-time elicitation strategies that unlock latent multi-view competence—task-specific chain-of-thought scaffolds and structured cross-view evidence aggregation—yielding substantial gains without retraining. 
Finally, we present preliminary evidence that reinforcement learning with verifiable rewards can elicit some latent multi-view competence in the base model, pointing to training-time approaches as a promising direction for future work.
Together, MVVBench offers a rigorous evaluation of 4D multi-view reasoning and a foundation for future progress toward reliable embodied perception.
\end{abstract}
\vspace{-0.5cm}

\section{Introduction}
\label{sec:intro}

Vision-Language Models (VLMs) have progressed rapidly from single-image understanding~\cite{liu2023visual,chen2022pali,liu2024improved} to multi-image and video comprehension~\cite{li2024llavaone,li2024llavanext,wang2024qwen2,google2025gemini3,hurst2024gpt,bai2025qwen3vltechnicalreport}. By encoding video as a sequence of frames, recent models have achieved strong results on temporal reasoning tasks including action recognition~\cite{li2024mvbench}, temporal grounding~\cite{ren2024timechat,liu2024tempcompass,cai2024temporalbench}, and long-form video QA~\cite{wang2024videoagent,fu2024video}. In parallel, multi-view image understanding has enabled static 3D awareness by jointly processing images from multiple viewpoints~\cite{li2024llavaone,zhu2025llava,hong20243dllm}.

However, real-world perception is inherently \emph{4D}: events unfold simultaneously across space and time, and an embodied agent must track entities as they transition between fields of view, synchronize actions across perspectives, and reason about latent scene continuity~\cite{shimojo2001visual,ogezi2025spare}. This requires jointly integrating \emph{spatial} evidence from multiple camera views and \emph{temporal} evidence across time---a capability that lies at the intersection of multi-view understanding and video reasoning, yet remains largely unexplored.

Existing benchmarks have probed these axes in isolation. Single-video benchmarks~\cite{li2024mvbench,liu2024tempcompass,fu2024video,videommev2_2026} test temporal reasoning but assume a single viewpoint. Multi-view image benchmarks~\cite{yeh2025seeing,chen2024spatialvlm,lee2025towards} evaluate spatial understanding from static snapshots but do not require temporal reasoning. Moreover, the few efforts addressing multi-view scene understanding~\cite{lee2025towards} have relied on VLM-generated questions, which upper-bounds the benchmark difficulty to the generator model's own comprehension and limits the ability to probe the frontiers of current systems. A benchmark that rigorously tests \emph{joint} multi-view and temporal reasoning---from questions that are provably unanswerable from any single view---is still missing.

We introduce \textbf{MVVBench}, a benchmark for multi-view video reasoning designed to fill this gap. MVVBench consists of 1{,}323 question--answer pairs drawn from 608 real-world multi-camera scenes (3{,}082 videos) sourced from Ego-Exo4D~\cite{grauman2024ego}, Panoptic Studio~\cite{Joo_2017_TPAMI}, and MMPTRACK~\cite{han2023mmptrack}. All questions are authored and verified by human experts under a strict \textbf{monocular-ambiguity} constraint: each question is unanswerable from any single view in the designated subset alone but becomes uniquely solvable by jointly reasoning over multiple views. This design naturally induces multi-hop reasoning and eliminates shortcuts available through single-view pattern matching. The benchmark spans six categories---implicit and explicit attribute identification, implicit and explicit relative distance, relative camera pose, and compositional counting---covering spatial reasoning, cross-view correspondence, and temporal dynamics.

Beyond benchmarking, we make four additional contributions. First, we conduct extensive diagnostic experiments that isolate the roles of visual evidence, temporal localization, and reasoning strategy, revealing that current VLMs struggle with cross-view entity correspondence and temporally coherent multi-view reasoning. 
Second, we design multi-view-specific chain-of-thought prompting strategies that outperform naive CoT, demonstrating that structured elicitation of multi-view reasoning yields substantial gains without retraining.
Third, we propose Tool-Augmented Dynamic Scene Graph (TADSG), an inference-time strategy that augments VLMs with off-the-shelf detection and re-identification tools to establish explicit cross-view identity, then constrains reasoning within a query-conditioned dynamic scene graph.  
Fourth, as a preliminary proof of concept, we explore reinforcement learning with verifiable rewards as a training-time complement to these inference-time strategies.
Together, MVVBench and our analysis provide a concrete path toward reliable 4D multi-view perception for embodied and real-world applications.

\section{Related Work}
\label{sec:related_work}

\paragraph{VLMs for video and 3D understanding.}
Early VLMs were designed for single-image comprehension~\cite{liu2023visual,chen2022pali,liu2024improved} and were subsequently extended to multi-image~\cite{li2024llavaone,li2024llavanext,awadalla2023openflamingo} and video inputs~\cite{lin2024video,zhang2025videollama,ren2024timechat}. Frontier models including Gemini~\cite{google2025gemini3,comanici2025gemini}, GPT~\cite{hurst2024gpt}, and the Qwen-VL family~\cite{qwen35,bai2025qwen3vltechnicalreport} now accept interleaved image and video streams and achieve strong performance on standard video QA. A parallel line of work targets 3D and multi-view spatial reasoning: 3D-LLM~\cite{hong20243dllm} and LLaVA-3D~\cite{zhu2025llava} inject 3D representations into LLMs, SpatialVLM~\cite{chen2024spatialvlm} trains for metric spatial relations from single images, and Cambrian-S~\cite{yang2025cambrian} augments VLMs with spatial sensing for video. These systems largely target either single-viewpoint temporal reasoning or static 3D awareness, leaving the integration of evidence across \emph{concurrent} dynamic camera streams underexplored.

\paragraph{VLM benchmarks.}
Single-video benchmarks have driven progress on temporal grounding, action ordering, and long-form comprehension, including MVBench~\cite{li2024mvbench}, TempCompass~\cite{liu2024tempcompass}, TemporalBench~\cite{cai2024temporalbench}, and Video-MME~\cite{fu2024video}. Most recently, MINERVA~\cite{nagrani2025minerva} targets complex multi-step video reasoning with hand-crafted questions, and Video-MME~v2~\cite{videommev2_2026} extends the Video-MME protocol into a progressive, hierarchically annotated suite, both pushing evaluation toward more challenging temporal reasoning regimes. Multi-view evaluation remains comparatively sparse and has so far been confined to static scenes. All-Angles Bench~\cite{yeh2025seeing} evaluates multi-view correspondence and geometric consistency over static image sets, while E3VQA~\cite{lee2025towards} provides QA pairs over static ego-exo frame pairs drawn from Ego-Exo4D. Notably, both benchmarks generate their QA pairs with VLMs (with light human filtering), which upper-bounds the achievable difficulty to the generator's own competence~-- a limitation we explicitly address by having human experts author every question under a monocular-ambiguity constraint. MVVBench differs from all of the above by jointly requiring dynamic temporal reasoning and cross-view spatial integration over concurrent multi-view \emph{videos}.

\paragraph{Reasoning strategies for VLMs.}
Chain-of-thought (CoT) prompting~\cite{wei2022chain,kojima2022large} has been widely adopted to elicit multi-step reasoning in both language and vision-language models~\cite{shao2024visual}. Beyond prompt-level reasoning, tool-augmented approaches~\cite{wang2024videoagent,liu2024llava} couple VLMs with external perception modules to offload grounding and tracking, and scene graph representations have been explored for spatial reasoning in embodied QA~\cite{saxena2024grapheqa} and video understanding~\cite{ji2020action,rosinol20203d}. However, generic CoT and tool-use strategies do not address the specific failure modes of multi-view video reasoning, namely cross-view entity correspondence and temporally coherent reasoning across concurrent streams. Our multi-view-guided CoT and Tool-Augmented Dynamic Scene Graph instantiate both directions for this setting, providing explicit cross-view identity and query-conditioned temporal scaffolding without retraining.

\begin{figure}
    \centering
    \includegraphics[width=\textwidth]{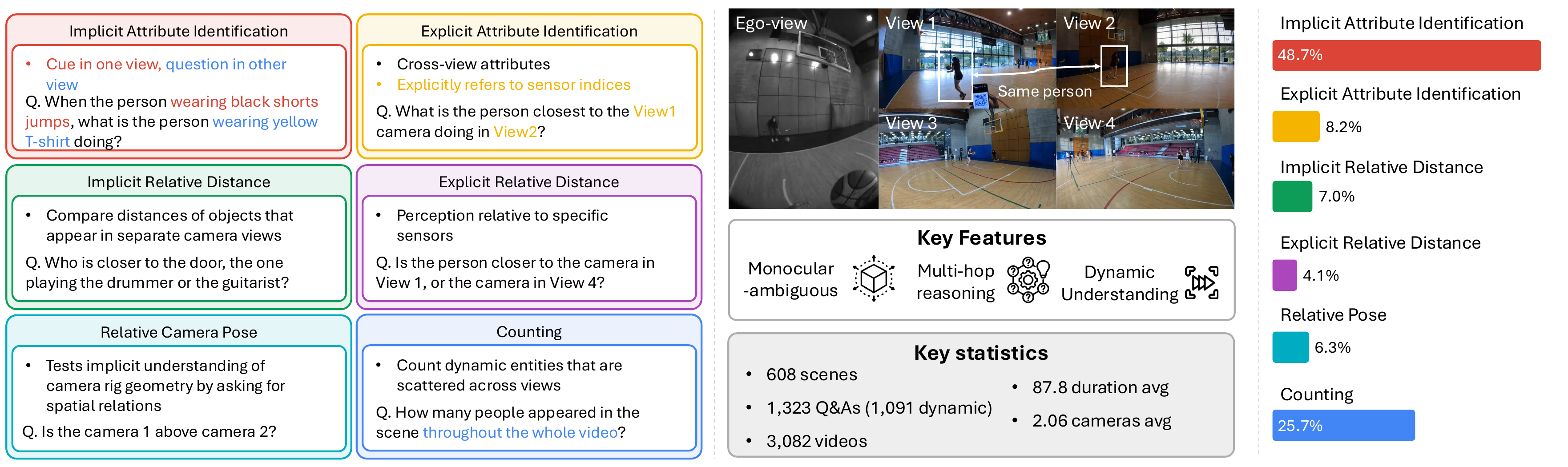}
    \caption{\textbf{Overview of MVVBench}. The benchmark consists of 1,323 Q\&A pairs from 608 scenes and 3,082 videos. MVVBench consists of 6 categories that tests understanding for different attributes, spatial understanding, and enumeration. All questions were generated by human experts focusing on ones that are monocular-ambiguous, requires multi-hop reasoning, and requires dynamic understanding of the 4D scene.}
    \label{fig:mvvbench_main}
    \vspace{-0.5cm}
\end{figure}

\section{MVVBench}
\label{sec:mvvbench}

\subsection{Data Collection}

\paragraph{Data sourcing.}
We source MVVBench from three multi-view video datasets: Ego-Exo4D~\cite{grauman2024ego} (496 scenes), Panoptic Studio~\cite{Joo_2017_TPAMI} (44 scenes), and MMPTRACK~\cite{han2023mmptrack} (68 scenes), totaling 608 scenes and 3{,}082 videos of diverse real-world activities. Ego-Exo4D was chosen for providing ego+exo pairs, Panoptic Studio was chosen for dense multi-view videos, and MMPTRACK provides surveillance-style videos, which provide rich diversity of possible multi-view configurations.

\paragraph{Annotation methodology.}
A common approach to benchmark construction is to generate QA pairs with a VLM, filter them with a second VLM, and verify with human annotators~\cite{yeh2025seeing,lee2025towards}. This paradigm inherently upper-bounds the difficulty of the benchmark to the generator model's own comprehension of the scene—a methodologically circular limitation when the benchmark's purpose is to probe the frontiers of current systems. In our preliminary experiments, we confirmed that VLM-generated questions rarely satisfied both factual correctness and the complexity required for rigorous 4D scene understanding. 

We therefore invert the conventional pipeline: (1)~\emph{human} QA generation, (2)~\emph{human} verification with iterative feedback, and (3)~VLM-assisted formatting. Three expert annotators with an average of three years of experience in computer vision annotation produced all QA pairs over a two-month period. 
The generated QAs were open-ended, and was aided with a minimal relevant temporal segment that should be seen in order to derive the answer.
A strict constraint was imposed: every question must be monocular-ambiguous, meaning the answer cannot be derived from any single video feed in isolation but is uniquely determinable when integrating information across views. This constraint naturally induces multi-hop reasoning, as the model must first ground the condition in one view and then reason about the question in another.

\paragraph{Task categories.}
We define six evaluation categories that challenge models to integrate spatial reasoning, cross-view correspondence, and temporal dynamics. Fig.~\ref{fig:mvvbench_main} provides an overview with representative examples. 
The categories probe two axes: \emph{attribute identification} and \emph{relative distance estimation}, each with an implicit variant (where the model must discover the relevant view) and an explicit variant (where the view index is given). In addition, \emph{relative camera pose} tests implicit understanding of the camera rig geometry, and \emph{counting} requires enumerating entities scattered across views, conditioned on attributes, actions, or temporal windows. We further focus on \emph{dynamic} QAs that are time-dependent or require temporal grounding; 1{,}091 out of 1{,}323 QAs fall into this category. Full definitions and additional examples are provided in Appendix \ref{appendix:mvvbench_details}.

\paragraph{Quality control.}
Throughout annotation, an expert verifier reviewed all QA pairs and provided iterative feedback whenever questions did not meet the benchmark's standards; 641 of the 1{,}323 QAs underwent at least one revision cycle. After all QAs were finalized, a second independent verifier validated the entire set for correctness and quality.

\begin{figure}
    \centering
    \includegraphics[width=\textwidth]{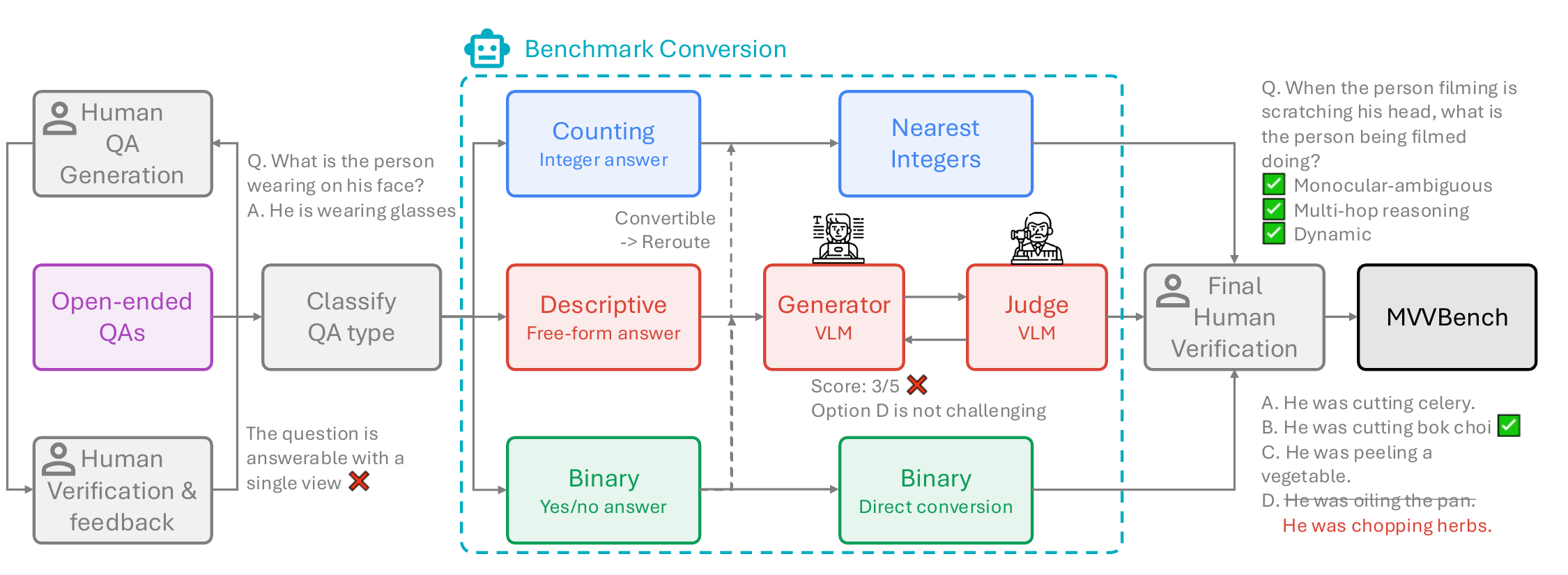}
    \caption{\textbf{Overview of MVVBench generation process}.
    \textit{(i) Human QA generation with iterative feedback}: expert annotators author open-ended QAs under a monocular-ambiguity constraint, with a verifier returning feedback until each question is monocular-ambiguous, multi-hop, and dynamic. 
    \textit{(ii) Automatic benchmark conversion (dashed box)}: QAs are classified by answer type. Counting QAs are converted via nearest-integer distractors; descriptive QAs pass through a generator–judge VLM loop that iterates until distractors are sufficiently challenging; binary QAs are either rerouted to the counting/descriptive branches or directly formatted as two-option MC. 
    \textit{(iii) Final human verification}: a final reviewer validates the converted item.}
    \label{fig:mvvbench_conversion}
    \vspace{-0.5cm}
\end{figure}

\vspace{-0.1cm}
\subsection{Benchmark Conversion}
\label{subsec:benchmark_conversion}

To facilitate reproducible and low-cost evaluation, we convert the open-ended QAs into a unified multiple-choice (MC) format via an automatic pipeline. See Fig.~\ref{fig:mvvbench_conversion} for an illustration.

\vspace{-0.1cm}
\paragraph{Answer-type classification.}
Each QA is first classified by its answer type: \emph{counting} (integer answer), \emph{descriptive} (free-form answer), or \emph{binary} (yes/no answer). Binary QAs receive additional processing: an LLM attempts to reframe each binary question as either a counting or descriptive question (e.g., ``Are there 3 people wearing white shoes?''\ becomes a counting question). Binary QAs that resist such reframing—typically those requiring multi-hop reasoning about cross-view attributes—are retained in their original form. 93 binary QAs were converted to either counting or descriptive QAs through this process.

\paragraph{Multiple-choice conversion.}
The conversion strategy is tailored to each answer type. Counting QAs are converted by selecting nearest-integer distractors, a process that requires no video access. Binary QAs are directly formatted as two-option MC questions. Descriptive QAs, however, demand careful distractor synthesis: wrong answers must be both unique and persuasive. Following VMCBench~\cite{zhang2025automated}, we employ a generator--judge loop with two VLM agents. The generator proposes distractors conditioned on the video, and the judge evaluates whether they are sufficiently challenging; if not, the generator produces a new set. A final round of human verification catches errors introduced during conversion. See Appendix~\ref{appendix:benchmark_conversion} for the details of the process.

\begin{table}[!t]
  \caption{\textbf{Main benchmark baseline results.} $\Delta$: Gain in overall score from lower
  bound.}
  \centering
  \small
  \setlength{\tabcolsep}{4.5pt}
  \renewcommand{\arraystretch}{1.15}
  \resizebox{\linewidth}{!}{%
  \begin{tabular}{@{} l
                  S[table-format=2.2] S[table-format=2.2]
                  S[table-format=2.2] S[table-format=2.2]
                  S[table-format=2.2] S[table-format=2.2]
                  S[table-format=2.2]
                  S[table-format=+2.2] @{}}
  \toprule
  Model &
  {Imp.\ Att.} & {Exp.\ Att.} & {Imp.\ D.} & {Exp.\ D.} &
  {Rel.\ Pose} & {Counting} & {\textbf{Overall}} & {$\Delta$} \\
  \midrule
  \rowcolor{gray!20}
  Random guess (lower bound) &
  41.50 & 32.64 & 49.70 & 47.84 & 25.00 & 25.63 & 35.81 & 0.00 \\
  \addlinespace[2pt]

  \midrule
  \multicolumn{9}{c}{\textit{Open-source VLMs}} \\
  \midrule
  Qwen3-VL-4B          & 43.83 & 37.19 & \cellcolor{tabthird} 63.10 & \cellcolor{tabsecond} 58.02
  & 41.67 & 35.20 & 42.78 & +6.97 \\
  Qwen3-VL-8B          & 45.08 & \cellcolor{tabsecond} 41.32 & 59.52 & \cellcolor{tabsecond}
  58.02 & 40.00 & 33.24 & 42.78 & +6.97 \\
  Qwen3-VL-32B         & 46.15 & 38.02 & 57.14 & 49.38 & 37.50 & 37.71 & 43.24 & +7.43 \\
  Qwen3.5-4B           & 45.08 & 36.36 & 53.57 & 51.85 & 31.67 & 31.01 & 40.21 & +4.40 \\
  Qwen3.5-9B           & 47.41 & 35.54 & 47.62 & 48.15 & 35.00 & 31.56 & 40.97 & +5.16 \\
  Qwen3.5-27B          & 49.02 & \cellcolor{tabthird} 39.67 & 55.95 & 54.32 & 37.50 & 29.89 &
  42.71 & +6.90 \\
  Qwen3.5-35B-A3B      & 49.91 & 38.02 & 55.95 & \cellcolor{tabthird} 56.79 & 36.67 & 28.21 &
  42.55 & +6.75 \\
  InternVL3.5-4B       & \cellcolor{tabthird} 54.92 & 37.19 & 59.52 & 46.91 & 28.33 & 39.66 &
  \cellcolor{tabthird} 46.56 & +10.75 \\
  InternVL3.5-8B       & 51.70 & 37.19 & 58.33 & 43.21 & 37.50 & 40.50 & 45.96 & +10.15 \\
  InternVL3.5-14B      & 52.06 & \cellcolor{tabthird} 39.67 & \cellcolor{tabsecond} 64.29 & 50.62
  & 19.17 & 36.59 & 44.44 & +8.64 \\
  InternVL3.5-38B      & 53.67 & \cellcolor{tabthird} 39.67 & \cellcolor{tabfirst} 65.48 & 49.38
  & 34.17 & 35.75 & 46.26 & +10.46 \\
  Gemma3-4B            & 39.89 & 32.96 & 24.79 & 32.50 & \cellcolor{tabsecond} 59.52 &
  \cellcolor{tabfirst} 51.85 & 37.94 & +2.14 \\
  Gemma3-12B           & 42.75 & \cellcolor{tabthird} 39.67 & 28.10 & 29.17 &
  \cellcolor{tabthird} 54.76 & \cellcolor{tabthird} 46.91 & 40.36 & +4.56 \\
  Gemma3-27B           & 40.61 & 34.71 & 48.81 & 49.38 & 53.33 & 40.22 & 42.18 & +6.38 \\
  LLaVA-onevision-7B   & 47.58 & 37.99 & 33.06 & 33.33 & 52.38 & \cellcolor{tabthird} 46.91 &
  42.63 & +6.83 \\
  LLaVA-onevision-72B  & \cellcolor{tabsecond} 55.81 & \cellcolor{tabfirst} 42.98 &
  \cellcolor{tabfirst} 65.48 & \cellcolor{tabfirst} 60.49 & 48.33 & 39.11 & \cellcolor{tabfirst}
  50.34 & +14.54 \\
  VideoLLaMA3-7B       & 45.26 & 30.45 & 34.71 & 25.00 & \cellcolor{tabfirst} 60.71 &
  \cellcolor{tabsecond} 49.38 & 39.68 & +3.88 \\
  Video-LLaVA-7B       & 39.71 & 26.26 & 29.75 & 23.33 & 47.62 & 40.74 & 34.24 & -1.56 \\
  GLM-4.6V-Flash-9B    & 44.19 & 30.45 & 33.06 & 37.50 & 53.57 & \cellcolor{tabsecond} 49.38 &
  39.76 & +3.96 \\
  MiniCPM-o-2.6-8B     & \cellcolor{tabfirst} 57.07 & \cellcolor{tabsecond} 41.32 & 55.95 & 50.62
  & 30.83 & 41.90 & \cellcolor{tabsecond} 48.68 & +12.88 \\
  Cambrian-S-7B        & 45.80 & 34.71 & 58.33 & 50.62 & 35.00 & 29.61 & 40.51 & +4.71 \\
  \addlinespace[3pt]

  \midrule
  \multicolumn{9}{c}{\textit{Closed-source VLMs}} \\
  \midrule
  Gemini-2.5-flash     & 54.92 & 46.28 & \cellcolor{tabsecond} 70.24 & \cellcolor{tabsecond}
  62.96 & 39.17 & 32.68 & 48.15 & +12.35 \\
  Gemini-2.5-pro       & \cellcolor{tabthird} 60.82 & \cellcolor{tabthird} 47.11 &
  \cellcolor{tabfirst} 72.62 & \cellcolor{tabfirst} 74.07 & 53.33 & 35.20 & \cellcolor{tabsecond}
  53.51 & +17.71 \\
  Gemini-3.0-flash     & \cellcolor{tabfirst} 61.90 & 29.61 & 44.63 & 35.83 &
  \cellcolor{tabthird} 65.48 & \cellcolor{tabfirst} 65.43 & 49.66 & +13.86 \\
  Gemini-3.0-pro       & 51.88 & 27.65 & 38.84 & 37.50 & 53.57 & 49.38 & 42.78 & +6.98 \\
  Gemini-3.1-pro       & \cellcolor{tabthird} 60.82 & \cellcolor{tabsecond} 47.93 & 57.14 &
  \cellcolor{tabthird} 58.02 & 40.00 & 34.36 & \cellcolor{tabthird} 50.19 & +14.38 \\
  GPT-4.1-mini         & 50.27 & 38.84 & 64.29 & 51.85 & 40.83 & 36.59 & 45.65 & +9.85 \\
  GPT-4.1              & 54.56 & 37.19 & 64.29 & \cellcolor{tabsecond} 62.96 & 40.00 & 41.06 &
  49.13 & +13.32 \\
  GPT-5-nano           & 52.77 & 35.54 & \cellcolor{tabthird} 67.86 & 53.09 & 33.33 & 40.22 &
  47.01 & +11.21 \\
  GPT-5-mini           & \cellcolor{tabsecond} 61.36 & 43.30 & 52.07 & 39.17 &
  \cellcolor{tabsecond} 67.86 & \cellcolor{tabsecond} 64.20 & \cellcolor{tabfirst} 54.20 & +18.40
  \\
  GPT-5.2              & 50.81 & 36.31 & 41.32 & 30.83 & \cellcolor{tabfirst} 70.24 &
  \cellcolor{tabthird} 54.32 & 45.65 & +9.85 \\
  GPT-5.4-mini         & 45.44 & 33.06 & 59.52 & 55.56 & 40.00 & 41.62 & 44.29 & +8.48 \\
  GPT-5.4              & 56.71 & \cellcolor{tabfirst} 48.76 & 57.14 & \cellcolor{tabthird} 58.02
  & 35.83 & 40.22 & 49.74 & +13.93 \\
  \addlinespace[3pt]

  \midrule
  \rowcolor{blue!20}
  Human performance &
  85.87 & 74.38 & 86.90 & 90.12 & 58.33 & 77.09 & 80.27 & +44.47 \\
  \bottomrule
  \end{tabular}
  }%
  \label{tab:main}
  \vspace{-0.5cm}
\end{table}

\section{Experiments}
\label{sec:experiments}

\subsection{Experimental setup}
\label{subsec:exp_setup}

\paragraph{VLM Settings}
We evaluate 33 VLMs spanning a wide capability spectrum, covering different model families and parameter counts. For closed-source models, we include Gemini-family variants spanning the 2.5, 3.0, and 3.1 series~\cite{google2025gemini3,comanici2025gemini}, and GPT-family variants spanning GPT-4.1 and GPT-5 series models~\cite{openai_gpt52_2025,openai_gpt5mini_model_2026}. For open-source models, we select from widely adopted and state-of-the-art model families: Qwen3-VL (4B/8B/32B)~\cite{bai2025qwen3vltechnicalreport}, 
Qwen3.5 (4B/9B/27B/35B-A3B)~\cite{qwen35},
InternVL3.5 (4B/8B/14B/38B)~\cite{wang2025internvl3}, Gemma3 (4B/12B/27B)~\cite{team2025gemma}, LLaVA-OneVision (7B/72B)~\cite{li2024llavaone}, VideoLLaMA3-7B~\cite{zhang2025videollama}, Video-LLaVA-7B~\cite{lin2024video}, GLM-4.6V-Flash-9B~\cite{vteam2025glm45vglm41vthinkingversatilemultimodal}, MiniCPM-o-2.6-8B~\cite{yao2024minicpm}, and Cambrian-S-7B~\cite{yang2025cambrian}.
Additionally, we report a random-guessing baseline that accounts for the benchmark’s mixture of binary and four-option multiple-choice questions, providing a calibrated lower bound. All experiments were run on the RTX PRO 6000 node with 96GB of VRAM per GPU. See Appendix~\ref{appendix:exp_details} for details in the prompts used.

\paragraph{Human Performance}
We report a human upper bound to contextualize VLM performance. We recruited an evaluator with one year of experience on computer-vision tasks to solve the full benchmark over one week. The evaluator was instructed to take as much time as needed for each problem, and we recorded the time spent per question. On average, answering a single question took 60.34 seconds, suggesting that the reasoning required is nontrivial even for humans. Details are provided in Appendix~\ref{appendix:human_perf_eval}.

\subsection{Main Results}
\label{subsec:main_results}

\begin{finding}
\label{find:chance}
Many VLMs remain near chance, and all models are far below human performance.
\end{finding}

Table~\ref{tab:main} shows that many VLMs operate only slightly above the chance-level lower bound (35.8), with Video-LLaVA-7B even falling below it. The spread between the weakest and strongest model is nearly 20 points, but the absolute ceiling among evaluated systems is GPT-5-mini at 54.2, which still trails human performance (80.3) by over 25 points; the best open-source model, LLaVA-OneVision-72B (50.3), trails by over 30. Across both open- and closed-source families, no model exceeds 55, indicating that MVVBench poses a genuine reasoning challenge beyond broad prior competence rather than a gap that current scale and training recipes are close to closing. Category-level inspection sharpens this picture: counting is the hardest axis overall—humans reach 90.1 while no VLM exceeds 65.4—and explicit relative distance is similarly difficult, with most models hovering near or below chance. Both categories require enumerating or comparing entities that are distributed across views, precisely the capability MVVBench is designed to probe.

\begin{finding}
\label{find:variance_meaningful}
MVVBench reveals highly non-monotonic and difficult-to-predict behavior across models, suggesting that it measures capabilities not captured by existing benchmark trends.
\end{finding}

Performance varies sharply both across families and within individual models across categories. For example, InternVL3.5-14B excels on implicit relative depth but collapses on relative pose, while Gemini-3.0-flash is strong on counting but weak on explicit relative depth. Scaling is non-monotonic even within the same family: larger variants do not consistently outperform smaller ones. If MVVBench scores were predictable from model family, scale, or existing leaderboard status~\cite{papailiopoulos2026benchpress}, the benchmark would add less value to the existing benchmarks. Instead, its irregular patterns suggest it probes partially independent capabilities---cross-view consistency, view-dependent grounding, multi-view aggregation---not well captured by standard evaluation suites.

\begin{table}[!t]
\caption{\textbf{Diagnostic experiments on Qwen3-VL-4B.} Text Only: no visual input; All Videos: full set of videos; Temp.\ Loc.: temporal localization provided. Qwen3-VL-4B-Thinking is included as a reasoning-model reference.}
\centering
\small
\setlength{\tabcolsep}{4.5pt}
\renewcommand{\arraystretch}{1.15}
\resizebox{\linewidth}{!}{%
\begin{tabular}{@{} l c
              S[table-format=2.2] S[table-format=2.2]
              S[table-format=2.2] S[table-format=2.2]
              S[table-format=2.2] S[table-format=2.2]
              S[table-format=2.2] @{}}
\toprule
Model & Method &
{Imp.\ Att.} & {Exp.\ Att.} & {Imp.\ D.} & {Exp.\ D.} &
{Rel.\ Pose} & {Counting} & {\textbf{Overall}} \\
\midrule

\rowcolor{gray!20}
Random guess (lower bound) &
- & 41.50 & 32.64 & 49.70 & 47.84 & 25.00 & 25.63 & 35.81 \\
\midrule

\multirow{4}{*}{Qwen3-VL-4B-Instruct}
& Text Only
& 39.36 & 27.27 & 57.14 \cellcolor{tabsecond} & 46.91
& 44.17 \cellcolor{tabthird} & 29.05 & 37.49 \\
& Baseline
& 43.83 \cellcolor{tabsecond} & 37.19 \cellcolor{tabthird} & 63.10 \cellcolor{tabfirst}
& 58.02 \cellcolor{tabsecond} & 41.67 & 35.20 \cellcolor{tabthird}
& 42.78 \cellcolor{tabthird} \\
& All Videos
& 44.72 \cellcolor{tabfirst} & 35.54 & 55.95 \cellcolor{tabthird}
& 55.56 \cellcolor{tabthird} & 45.00 \cellcolor{tabsecond}
& 37.43 \cellcolor{tabfirst} & 43.31 \cellcolor{tabsecond} \\
& Temp.\ Loc.
& 44.72 \cellcolor{tabfirst} & 41.32 \cellcolor{tabfirst} & 55.95 \cellcolor{tabthird}
& 59.26 \cellcolor{tabfirst} & 45.83 \cellcolor{tabfirst}
& 35.75 \cellcolor{tabsecond} & 43.69 \cellcolor{tabfirst} \\
\midrule

Qwen3-VL-4B-Thinking & Baseline
& 42.93 \cellcolor{tabthird} & 38.84 \cellcolor{tabsecond} & 51.90
& 48.15 & 41.86 & 33.24 & 40.65 \\
\bottomrule
\end{tabular}
}%
\label{tab:analyzing_and_improving}
\vspace{-0.5cm}
\end{table}

\subsection{Diagnostic Analysis}
\label{subsec:diagnostic}

To better understand what drives success on MVVBench, we perform a series of controlled experiments centered on Qwen3-VL-4B-Instruct, while also including Qwen3-VL-4B-Thinking as a reference point. Table~\ref{tab:analyzing_and_improving} summarizes the results. These experiments are designed to separate several factors that are otherwise entangled in the main benchmark: language prior exploitation, access to extra visual evidence, temporal localization, and reasoning strategy.

\begin{finding}
\label{find:text_only_not_enough}
Text-only performance remains close to chance, showing that MVVBench cannot be solved from linguistic priors alone.
\end{finding}

A natural concern for any multimodal benchmark is whether questions can be answered from textual regularities alone~\cite{tong2024eyes}. In a ``text only'' setting (question and choices, no visual input), the model scores 37.49---only marginally above the random-guess baseline of 35.81 and far below the visual baseline of 42.78. This confirms that MVVBench meaningfully requires visual evidence and is relatively robust to hacking through text priors.

\begin{finding}
\label{find:extra_view_cheating}
Providing access to all videos inflates performance by allowing the model to exploit visual evidence outside the designated evaluation views.
\end{finding}

In the ``all videos'' setting, the model receives the full set of videos rather than the selected views. This raises the Overall score modestly from 42.78 to 43.31, confirming that additional views can serve as a shortcut. The monocular-ambiguity constraint is enforced \emph{only over the selected views}: by construction, no single view in the selected set reveals the answer in isolation. When the model is given access to the full scene, however, an unselected view may directly expose the answer, allowing the model to bypass cross-view reasoning entirely and answer from a single view. MVVBench is therefore designed to test whether the model can reason from a \emph{constrained} set of views.

\begin{finding}
\label{find:temporal_localization}
Temporal localization yields a clear improvement, indicating that event retrieval is a substantial bottleneck in MVVBench.
\end{finding}

Recall that MVVBench also contains the relevant time segment marked by the annotators.
Providing the ground-truth temporal window (``Temporal Localization'') raises the Overall score to 43.69, outperforming both the baseline and the all-videos setting, with notable gains on explicit attribute identification and explicit relative distance. This shows that temporal localization is itself a nontrivial subproblem~\cite{li2024mvbench,liu2024tempcompass,cai2024temporalbench}: models struggle not only with cross-view aggregation but also with identifying \emph{when} the relevant event occurs. 
Because every MVVBench question is annotated with the exact temporal window required to answer it, this observation points to two additional uses of the benchmark beyond its default protocol. First, MVVBench can serve as a testbed for multi-view temporal grounding in its own right—evaluating whether a model can recover the annotated window from the full video, a direction we leave to future work. Second, for studies that aim to isolate multi-view reasoning from event retrieval, the annotated windows provide a clean way to disentangle the two: feeding only the relevant segment removes temporal localization as a confounder and yields a sharper lens on the remaining cross-view reasoning difficulty.

\begin{figure}
    \centering
    \includegraphics[width=\textwidth]{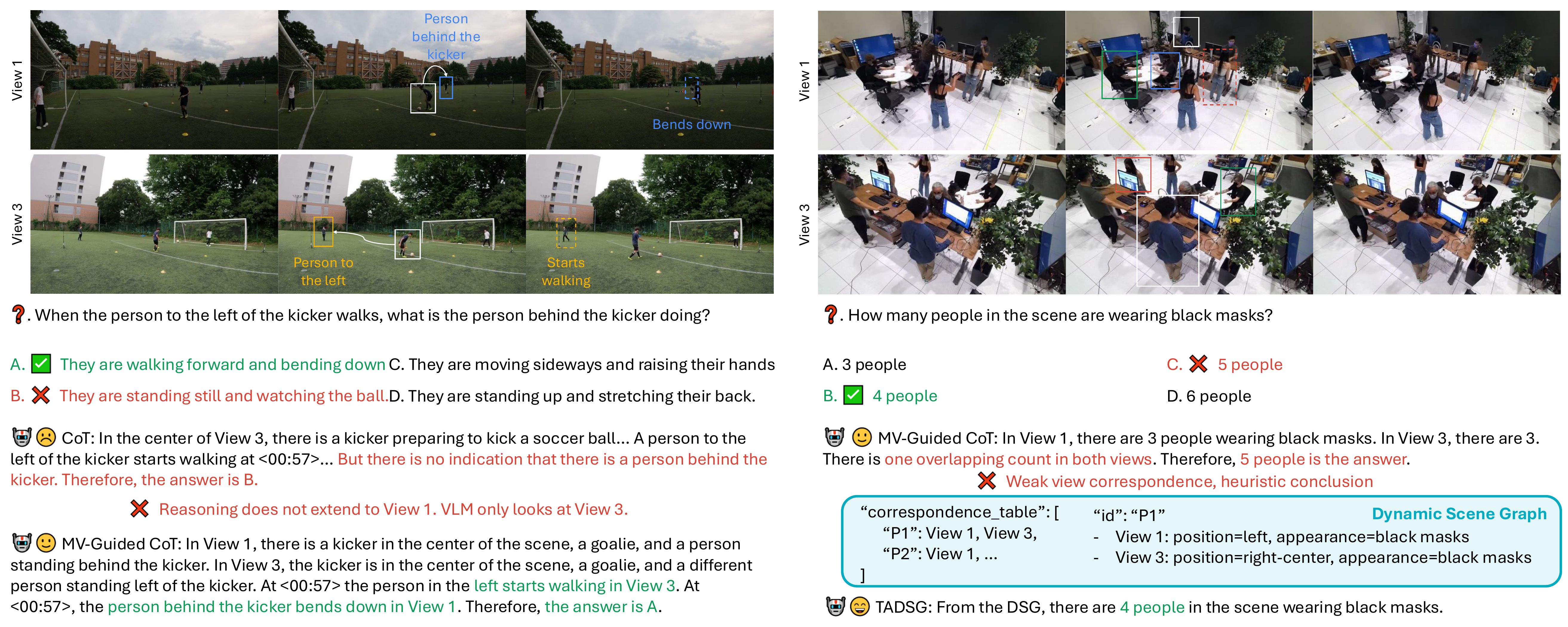}
    \caption{\textbf{Examples of MVVBench QAs and VLM reasoning failure modes.}
    \textit{(Left) - Implicit Attribute Identification}. 
    Failure mode of naive CoT. MV-Guided CoT corrects the reasoning process.
    \textit{(Right) - Counting}. 
    Failure mode of MV-Guided CoT. TADSG builds an explicit cross-view correspondence to answer correctly.}
    \label{fig:examples_and_failure}
    \vspace{-0.5cm}
\end{figure}

\subsection{Structured Reasoning for Multi-View Video}
\label{subsec:structured_reasoning}

Given the analysis in \S\ref{subsec:diagnostic}, we then aim to explore if inference-time reasoning strategies can improve the multi-view reasoning behavior of existing VLMs. We present the results in Tab.~\ref{tab:reasoning_static_dynamic}.

\begin{finding}
\label{find:naive_cot_not_helpful}
Naive chain-of-thought prompting does not improve multi-view reasoning. Performance improves when the reasoning process is explicitly structured for the multi-view setting.
\end{finding}

Naive CoT prompting~\cite{wei2022chain,kojima2022large} does not help: it slightly \emph{lowers} the Overall score from 42.78 to 42.55, and the dedicated thinking model also underperforms the instruct baseline (40.65 vs.\ 42.78).
We find that the degeneration often happens from the reasoning chain getting stuck on a single view or repeating itself, as can be seen in the example of Fig.~\ref{fig:examples_and_failure}. To mitigate this, we introduce MV-Guided CoT, which explicitly guides the reasoning process to sequentially reason over the views separately, and combine the information by cross referencing, while emphasizing that each view shows the same scene rendered from different views.
See Tab.~\ref{tab:prompt_cot} to see the difference in the prompts.
With this modification, MV-Guided CoT raises the overall accuracy to 44.82. This confirms that multi-view reasoning requires structured elicitation rather than simply longer reasoning traces.
Despite these gains, inspecting the remaining failures reveals two recurring bottlenecks that prompting alone cannot resolve.

\begin{finding}
\label{find:weak_view_correspondence}
Even under explicit view-by-view reasoning, models do not reliably establish correspondence across views.
\end{finding}

A central failure mode is weak view correspondence: models reason about each view independently but fail to verify which entity in one view matches which entity in another. This leads to brittle heuristics—e.g., describing people separately per view and applying ad hoc duplicate removal without grounded identity checks. Fig.~\ref{fig:examples_and_failure} (right) shows a representative case: MV-Guided CoT counts people wearing the target attribute within each view (3 and 3), then subtracts a single guessed overlap to arrive at 5, missing the true count of 4. The error is not one of perception within any view but of cross-view identity resolution. While MV-Guided CoT encourages view-by-view inspection, it still lacks a reliable mechanism for cross-view entity linking—precisely the gap that motivates our proposed explicit correspondence stage in \S\ref{sec:tadsg}.

\begin{finding}
\label{find:dynamic_questions_harder}
Failures become substantially worse on dynamic questions, where temporal reasoning and cross-view alignment must be maintained over time.
\end{finding}

This problem is amplified for dynamic questions, where the model must additionally maintain temporally coherent representations---\emph{who} is being tracked, \emph{when} the event occurs, and \emph{how} observations from different views refer to the same evolving scene. Tab.~\ref{tab:reasoning_static_dynamic} confirms that all methods degrade on dynamic questions relative to static ones. These bottlenecks motivate an inference-time strategy that provides explicit cross-view identity support.

\newcolumntype{M}{>{\raggedright\arraybackslash}m{2.05cm}}

\newcommand{\gapdown}[1]{\textcolor{red!70!black}{\scriptsize($\downarrow$#1)}}
\newcommand{\gapup}[1]{\textcolor{green!50!black}{\scriptsize($\uparrow$#1)}}
\newcommand{\gapeq}{\textcolor{black!60}{\scriptsize($\rightarrow$0.00)}}

\newcommand{\metric}[1]{%
  \hspace*{0.25em}%
  \makebox[3.2em][r]{#1}%
}

\newcommand{\metricdelta}[2]{%
  \metric{#1}%
  \hspace{0.08em}%
  #2%
}

\begin{table}[!t]
\caption{\textbf{Static vs.\ dynamic breakdown for multi-step reasoning methods on MVVBench.} Dynamic rows also show the point difference relative to the corresponding static accuracy.}
\centering
\small
\setlength{\tabcolsep}{1.0pt}
\renewcommand{\arraystretch}{1.03}
\resizebox{\linewidth}{!}{%
\begin{tabular}{@{} l c
                *{7}{M} @{}}
\toprule
Method & Split &
\multicolumn{1}{c}{Imp.\ Att.} &
\multicolumn{1}{c}{Exp.\ Att.} &
\multicolumn{1}{c}{Imp.\ D.} &
\multicolumn{1}{c}{Exp.\ D.} &
\multicolumn{1}{c}{Rel.\ Pose} &
\multicolumn{1}{c}{Counting} &
\multicolumn{1}{c}{\textbf{Overall}} \\
\midrule

\multirow{3}{*}{CoT~\cite{kojima2022large}}
& Static
& \metric{59.80}
& \metric{66.67}
& \metric{64.71}
& \metric{60.00}
& \metric{45.83}
& \metric{36.54}
& \metric{52.16} \\
& Dynamic
& \metricdelta{44.86}{\gapdown{14.94}}
& \metricdelta{26.27}{\gapdown{40.40}}
& \metricdelta{59.70}{\gapdown{5.01}}
& \metricdelta{46.48}{\gapdown{13.52}}
& \metricdelta{43.06}{\gapdown{2.77}}
& \metricdelta{33.33}{\gapdown{3.21}}
& \metricdelta{40.51}{\gapdown{11.65}} \\
& Overall
& \cellcolor{tabthird}\metric{47.58}
& \metric{27.27}
& \metric{60.71}
& \metric{48.15}
& \cellcolor{tabthird}\metric{44.17}
& \cellcolor{tabsecond}\metric{33.80}
& \metric{42.55} \\
\cmidrule(lr){1-9}

\multirow{3}{*}{MV-Guided CoT (\textbf{ours})}
& Static
& \metric{52.94}
& \metric{33.33}
& \metric{64.71}
& \metric{50.00}
& \metric{50.00}
& \metric{32.69}
& \metric{48.28} \\
& Dynamic
& \metricdelta{44.42}{\gapdown{8.52}}
& \metricdelta{48.31}{\gapup{14.98}}
& \metricdelta{68.66}{\gapup{3.95}}
& \metricdelta{52.11}{\gapup{2.11}}
& \metricdelta{51.39}{\gapup{1.39}}
& \metricdelta{33.01}{\gapup{0.32}}
& \metricdelta{44.09}{\gapdown{4.19}} \\
& Overall
& \metric{45.97}
& \cellcolor{tabfirst}\metric{47.93}
& \cellcolor{tabsecond}\metric{67.86}
& \cellcolor{tabthird}\metric{51.85}
& \cellcolor{tabfirst}\metric{50.83}
& \cellcolor{tabthird}\metric{32.96}
& \cellcolor{tabthird}\metric{44.82} \\
\cmidrule(lr){1-9}

\multirow{3}{*}{M3CoT~\cite{lee2025towards}}
& Static
& \metric{51.96}
& \metric{66.67}
& \metric{70.59}
& \metric{60.00}
& \metric{50.00}
& \metric{34.62}
& \metric{49.57} \\
& Dynamic
& \metricdelta{47.05}{\gapdown{4.91}}
& \metricdelta{40.68}{\gapdown{25.99}}
& \metricdelta{68.66}{\gapdown{1.93}}
& \metricdelta{52.11}{\gapdown{7.89}}
& \metricdelta{50.00}{\gapeq}
& \metricdelta{33.66}{\gapdown{0.96}}
& \metricdelta{44.45}{\gapdown{5.12}} \\
& Overall
& \cellcolor{tabsecond}\metric{47.94}
& \cellcolor{tabsecond}\metric{41.32}
& \cellcolor{tabfirst}\metric{69.04}
& \cellcolor{tabsecond}\metric{53.08}
& \cellcolor{tabsecond}\metric{50.00}
& \cellcolor{tabsecond}\metric{33.80}
& \cellcolor{tabsecond}\metric{45.35} \\
\cmidrule(lr){1-9}

\multirow{3}{*}{TADSG (\textbf{ours})}
& Static
& \metric{54.90}
& \metric{66.67}
& \metric{52.94}
& \metric{50.00}
& \metric{52.08}
& \metric{36.54}
& \metric{50.00} \\
& Dynamic
& \metricdelta{49.02}{\gapdown{5.88}}
& \metricdelta{38.14}{\gapdown{28.53}}
& \metricdelta{70.15}{\gapup{17.21}}
& \metricdelta{57.75}{\gapup{7.75}}
& \metricdelta{50.00}{\gapdown{2.08}}
& \metricdelta{39.22}{\gapup{2.68}}
& \metricdelta{47.02}{\gapdown{2.98}} \\
& Overall
& \cellcolor{tabfirst}\metric{50.09}
& \cellcolor{tabthird}\metric{38.84}
& \cellcolor{tabthird}\metric{66.67}
& \cellcolor{tabfirst}\metric{56.79}
& \cellcolor{tabfirst}\metric{50.83}
& \cellcolor{tabfirst}\metric{38.83}
& \cellcolor{tabfirst}\metric{47.54} \\

\bottomrule
\end{tabular}
}%
\label{tab:reasoning_static_dynamic}
\vspace{-0.5cm}
\end{table}

\subsection{Enhancing the performance in MVVBench}

\paragraph{Tool-Augmented Dynamic Scene Graph}
\label{sec:tadsg}

The failure analysis identifies two core bottlenecks: weak cross-view entity correspondence and the inability to maintain temporally coherent reasoning across views. We propose Tool-Augmented Dynamic Scene Graph (TADSG), an inference-time strategy that addresses both by (1) augmenting the VLM with off-the-shelf vision tools for cross-view identity establishment, and (2) constraining the reasoning process within a query-conditioned dynamic scene graph (DSG)~\cite{rosinol20203d,ji2020action}.

TADSG operates in three stages. First, a person detector~\cite{yolov8_ultralytics} and a re-identification model (ReID)~\cite{zhou2019osnet} are used to detect people across views and cluster them into global identities via cross-view appearance matching---providing the explicit correspondence that current VLMs cannot reliably infer from pixels alone. As shown in Fig.~\ref{fig:examples_and_failure} (Right), the tools generate a correspondence table, indicating which entities are seen from each view, along with matches.
Second, the VLM resolves the question's natural-language reference to a global identity and constructs a target-centric DSG: a structured representation containing per-view anchor and temporal observations, cross-view relations, and question-relevant facts for the queried entity only. 
Third, the VLM answers the question using the graph as its primary evidence rather than raw video.

A critical design choice is confidence gating: TADSG activates only when cross-view correspondence and target identity are sufficiently reliable, falling back to the prompting baseline otherwise. This selective activation ensures that gains come from high-confidence cases rather than from uniformly applying a potentially noisy graph. Full implementation details, the graph schema, and gating thresholds are provided in Appendix~\ref{appendix:tadsg}.

\paragraph{Toward training-time improvement via RL.}
The inference-time strategies above (MV-Guided CoT, TADSG) improve multi-view reasoning through external scaffolding, raising a natural question: can reinforcement learning \emph{distill} such structured reasoning directly into the base model? We conduct a preliminary study using DAPO~\cite{yu2025dapo} with a verifiable reward (exact-match on the correct option letter) to fine-tune Qwen3-VL-4B-Instruct on a subset of MVVBench. We filter 1{,}254 two-view questions across all categories, hold out 300 for evaluation, and train on the remaining 954 with LoRA. After post-training, we observe that the accuracy increases from 43.67\% to 47.67\%. Full details are in Appendix~\ref{appendix:rlvr}.

\section{Conclusion}
\label{sec:conclusion}

We introduced MVVBench, a benchmark for multi-view video reasoning consisting of 1{,}323 human-authored QA pairs across 608 real-world multi-camera scenes. By enforcing monocular ambiguity---every question is unanswerable from any single view within the designated input set---MVVBench isolates the ability to jointly integrate spatial evidence across views and temporal evidence across time, a capability not probed by existing benchmarks.
Our extensive evaluation of different VLMs reveals that most models perform only marginally above chance, with even the best model trailing human performance by over 25 points. Diagnostic experiments show that MVVBench is not solvable from textual priors, that temporal localization is a key bottleneck, and that naive chain-of-thought reasoning does not help. Gains emerge only when the reasoning process is explicitly structured for the multi-view setting: our MV-Guided CoT and TADSG methods yield consistent improvements by scaffolding cross-view correspondence and evidence aggregation.

\paragraph{Limitations.}
MVVBench currently focuses on human-centric scenes due to the source datasets, which limits generalization to scenarios involving other object types. The TADSG pipeline relies on person-specific detection and re-identification models, constraining its applicability to scenes with people. Furthermore, our benchmark contains a fixed set of six categories, which may not cover all aspects of multi-view reasoning such as fine-grained action understanding or causal reasoning across views. Future work should expand the domain coverage, explore training-time approaches such as reinforcement learning for multi-view reasoning, and develop more general cross-view correspondence methods.

\paragraph{Broader impacts.}
MVVBench is intended to support the development and evaluation of VLMs that can more reliably reason over dynamic multi-camera scenes. Potential positive impacts include improving embodied perception, assistive robotics, safety monitoring, and diagnostic analysis of VLM failure modes in settings where multi-view temporal reasoning is required. At the same time, stronger multi-view video understanding could be misused in surveillance or tracking applications, especially when combined with person detection or re-identification tools. We therefore emphasize that MVVBench is a research benchmark rather than a deployed monitoring system, and that responsible use should consider privacy, consent, dataset licensing, and the risks of identity tracking in real-world environments. Our tool-augmented analysis is used only for benchmark diagnosis and does not release a new person-tracking system.


\bibliographystyle{plain}
\bibliography{main}
\clearpage
\appendix

\section{Further Details on MVVBench}
\label{appendix:mvvbench_details}

\begin{figure}
    \centering
    \includegraphics[width=\textwidth]{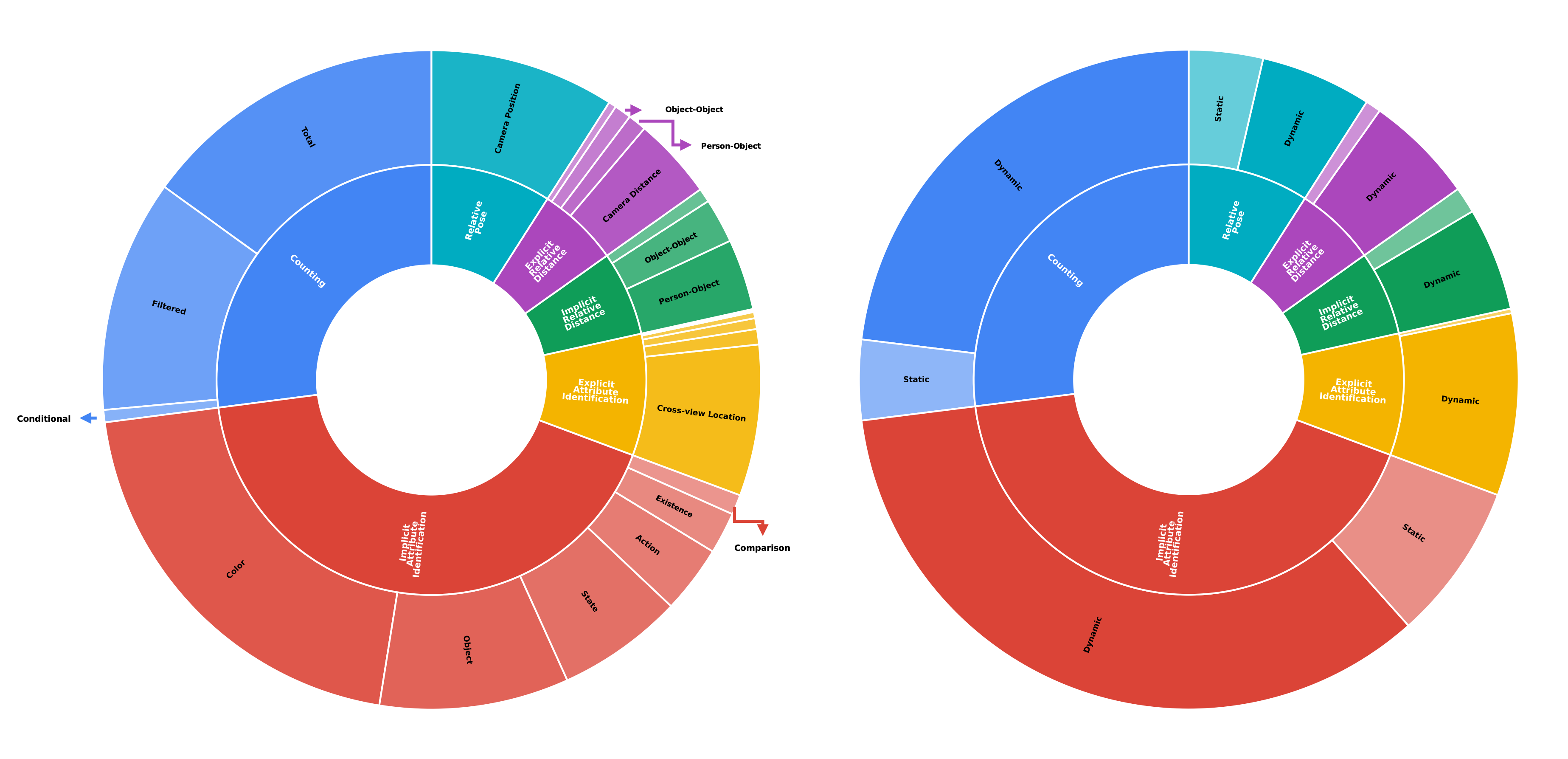}
    \caption{(Left): Distribution of MVVBench by different types of subcategories, (Right): Distribution of MVVBench by dynamic/static Q\&As.}
    \label{fig:mvvbench_subcategory}
\end{figure}

MVVBench contains the following six categories.

\begin{enumerate}
    \item \textbf{Implicit Attribute Identification}: This tests the model's ability to maintain identity and context across disjoint views. Questions may define a temporal or attribute cue in one view (e.g., "When the person in black jumps...") and request an attribute or action description from a different view (e.g., "...what is the person in yellow doing?").
    \item \textbf{Explicit Attribute Identification}: These questions explicitly reference the sensor indices, testing the model's understanding of camera proximity and spatial layout (e.g., "What is the person closest to the View 1 camera doing in View 2?").
    \item \textbf{Implicit Relative Distance}: This category assesses global spatial reasoning. It asks models to compare the distances of objects that may appear in separate camera views (e.g., comparing the distance of a person in View 1 to a door visible only in View 2).
    \item \textbf{Explicit Relative Distance}: This evaluates depth perception relative to specific sensors. Questions involve comparing distances between an object and two different cameras (e.g., "Is the person closer to the camera in View 1 or View 4?").
    \item \textbf{Relative Camera Pose}: This tests the model’s implicit understanding of the camera rig geometry by asking for spatial relationships between the sensors themselves (e.g., "Is the camera for View 4 located to the left or right of the camera for View 1?").
    \item \textbf{Counting}: Beyond simple enumeration, this category requires counting objects conditioned on specific attributes (e.g., color), actions (e.g., "climbing"), or temporal windows (e.g., "count people during the interval where the black-shirted person throws the ball").
\end{enumerate}

For the category visualization in Figure~\ref{fig:mvvbench_subcategory}, we further partitioned each original MVVBench question category into a small set of finer-grained subcategories using
a lightweight rule-based procedure. The goal of this taxonomy was descriptive rather than canonical. Accordingly, we kept the original MVVBench category as the inner-ring label in the visualization and assigned
each question a single outer-ring subcategory based on its linguistic form and intended reasoning type.

\paragraph{Counting.}
Counting questions were divided into \textit{Total}, \textit{Filtered}, and \textit{Conditional}. \textit{Total} corresponds to direct counting questions that ask for the
number of entities in the scene without any additional qualifier. \textit{Filtered} captures counting over a subset defined by an attribute or state (e.g., entities wearing
something, having a certain color, or satisfying some property). \textit{Conditional} captures questions where the count is tied to a temporal or situational clause, such as
counting before, after, during, or while some event occurs.

\paragraph{Attribute Identification.}
Attribute-identification questions were divided into \textit{Action}, \textit{Object}, \textit{Color}, \textit{Existence}, \textit{Comparison}, and \textit{State}.
\textit{Action} includes questions asking what a person is doing. \textit{Object} includes identifying what an entity is holding, wearing, carrying, taking out, and similar
object-centric relations. \textit{Color} captures direct color queries. \textit{Existence} groups yes/no verification questions about whether an object, part, or attribute is
present. \textit{Comparison} captures same/different style attribute comparisons. \textit{State} is a residual bucket for appearance or status questions that do not cleanly
fall into the previous groups.

For \textit{Attribute identification with view index}, we used the same logic but added a dedicated \textit{Cross-view Location} bucket for questions that explicitly ask
where an entity seen in one view is located in another view. In practice, this bucket accounts for most of the view-indexed attribute questions.

\paragraph{Relative Distance.}
Relative-distance questions were divided according to what kinds of entities are being compared: \textit{Person-Object}, \textit{Person-Person}, \textit{Object-Object}, and
\textit{Camera Distance}. The first three buckets are meant to distinguish whether the proximity judgment is anchored between people, between objects, or across a person/
object pair. For \textit{Relative distance with view index}, we additionally use \textit{Camera Distance} for questions that explicitly compare which camera or view is
closer/farther from a referenced object or region.

\paragraph{Relative Pose.}
Relative-pose questions were assigned to \textit{Camera Position} when the wording explicitly referred to view-to-view or camera-to-camera spatial relations. We also included
a fallback \textit{Spatial Relation} bucket for generic object-centric pose questions. However, in the current MVVBench release, essentially all questions in this category
matched the camera/view-relative pattern, so the populated subcategory is effectively \textit{Camera Position}.

\section{Labeling Tool}
\label{appendix:labeling_tool}

\begin{figure}
    \centering
    \includegraphics[width=\textwidth]{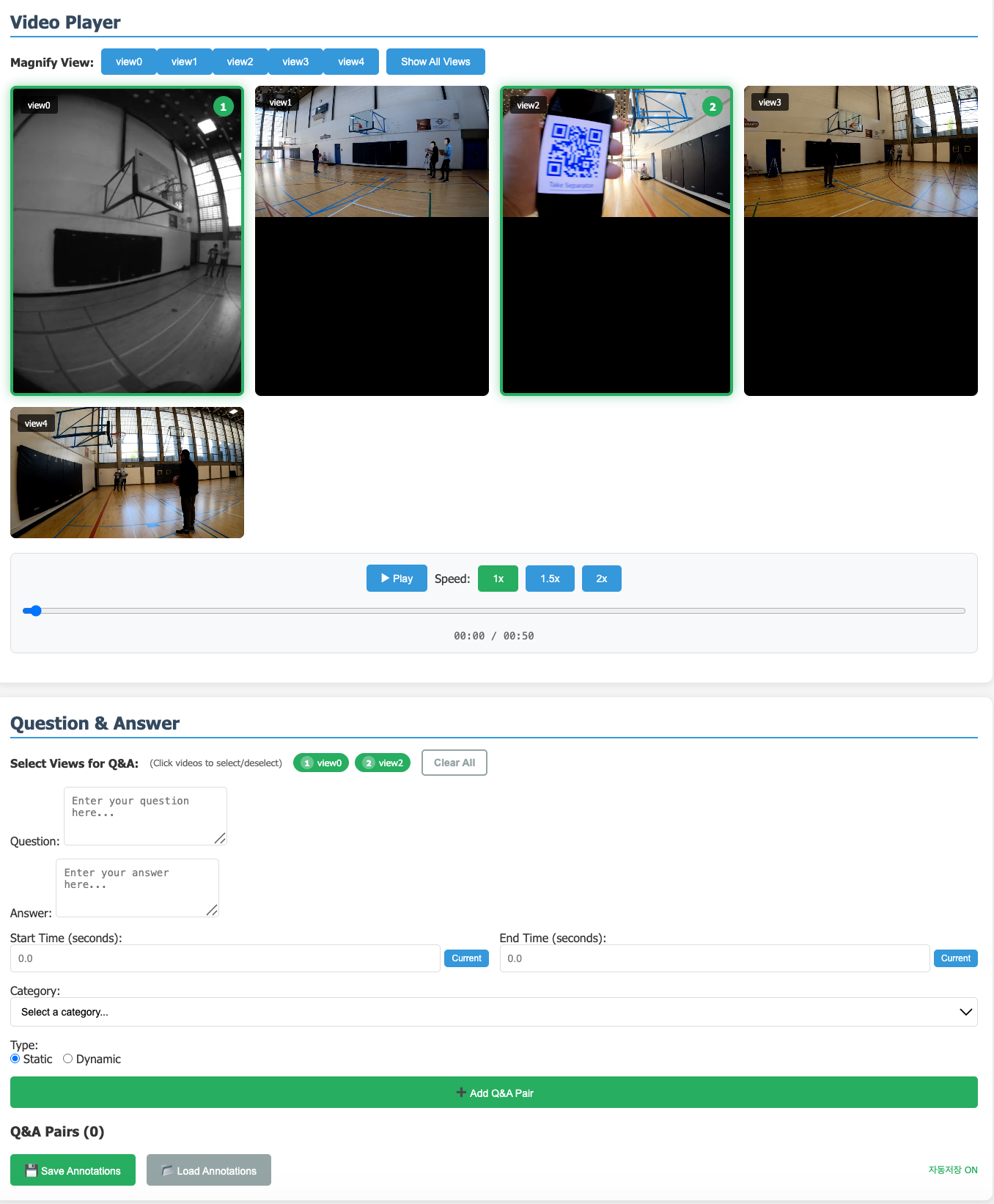}
    \caption{\textbf{Screenshot of the labeling tool (annotator side)}.}
    \label{fig:labeling_tool_annotator}
\end{figure}

\begin{figure}
    \centering
    \includegraphics[width=\textwidth]{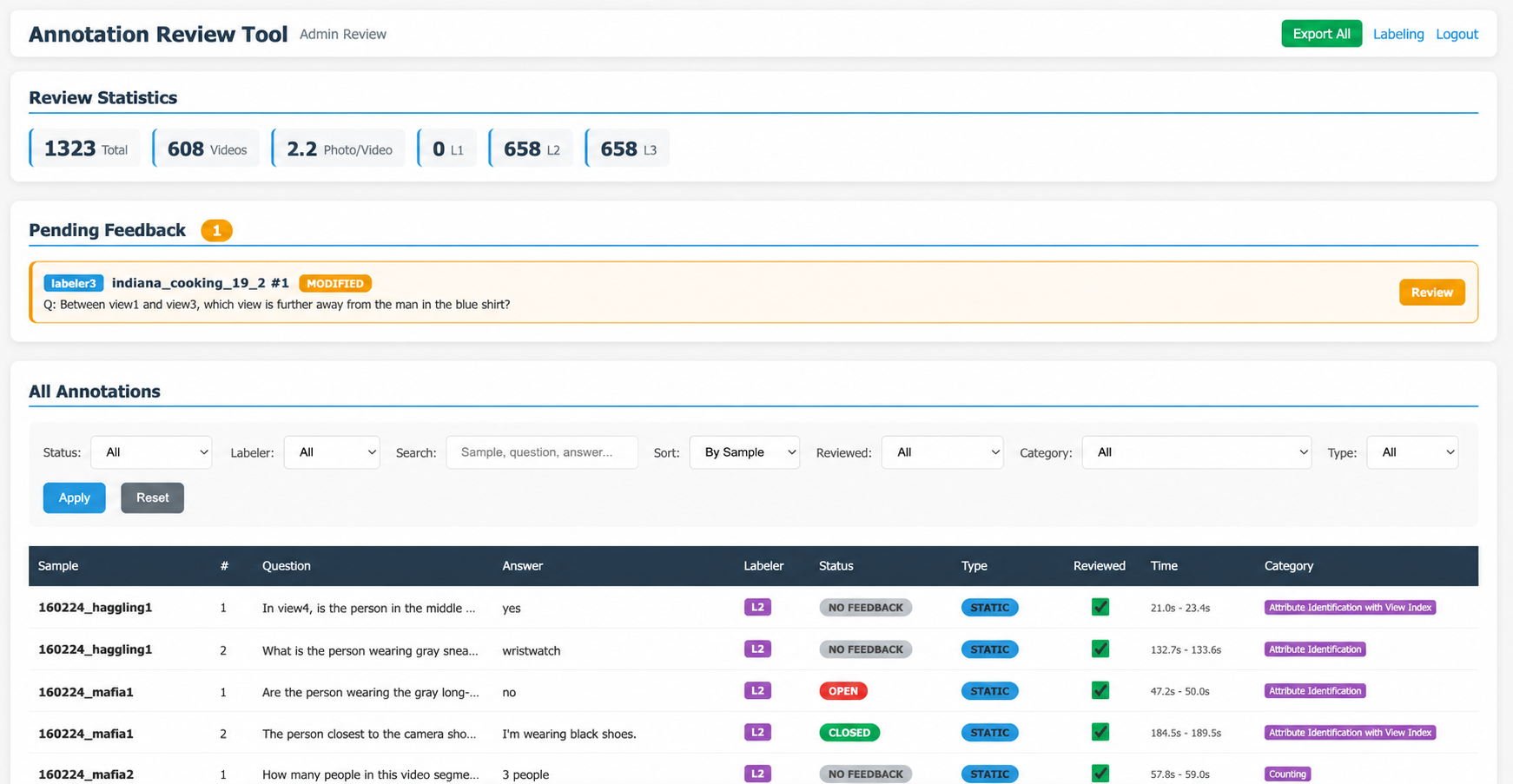}
    \caption{\textbf{Screenshot of the labeling tool (verifier side)}.}
    \label{fig:labeling_tool_admin0}
\end{figure}

\begin{figure}
    \centering
    \includegraphics[width=\textwidth]{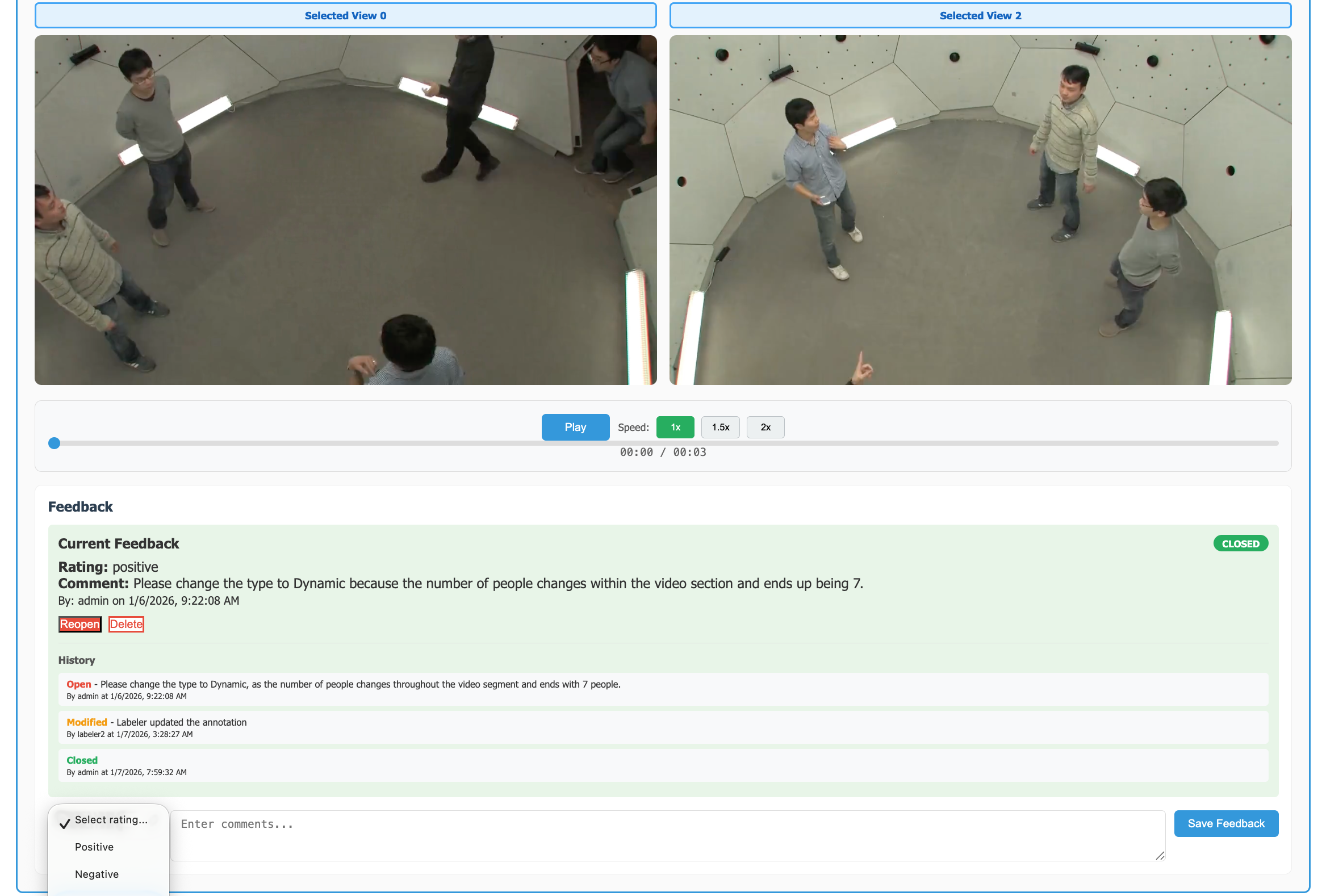}
    \caption{\textbf{Screenshot of the labeling tool (verifier side)}.}
    \label{fig:labeling_tool_admin1_feedback}
\end{figure}

We developed a custom web-based labeling tool that manages the data annotation workflow through secure, individual user authentication. Upon logging in, annotators are presented with a curated queue of video sequences uniquely assigned to their ID (see Fig.~\ref{fig:labeling_tool_annotator}). The interface supports synchronized multi-view playback with adjustable speed controls, allowing the annotator to analyze the scene from various angles. To generate a data point, the annotator selects the relevant viewpoints (typically two or more) and formulates a Question-Answer (QA) pair. Furthermore, they are required to classify the QA category and designate the specific temporal interval required to answer the question. The completed annotation is stored in JSON format.

Verifiers access the platform via administrator privileges to audit the dataset generated by the annotators. As shown in Fig.~\ref{fig:labeling_tool_admin0}, the administrative dashboard provides comprehensive search and filtering capabilities, allowing verifiers to query annotations by video name, category, or annotator ID. When reviewing a specific instance (Fig.~\ref{fig:labeling_tool_admin1_feedback}), the verifier assesses the quality of the QA pair. They may either validate the entry with a ``Positive'' rating or reject it with a ``Negative'' rating. If rejected, the verifier must provide constructive feedback to guide the revision process. These flagged instances are aggregated in the ``Pending Feedback'' section (Fig.~\ref{fig:labeling_tool_admin0}), which remains accessible to both the verifier and the respective annotator to ensure an iterative quality control loop.

\begin{figure}
    \centering
    \includegraphics[width=\textwidth]{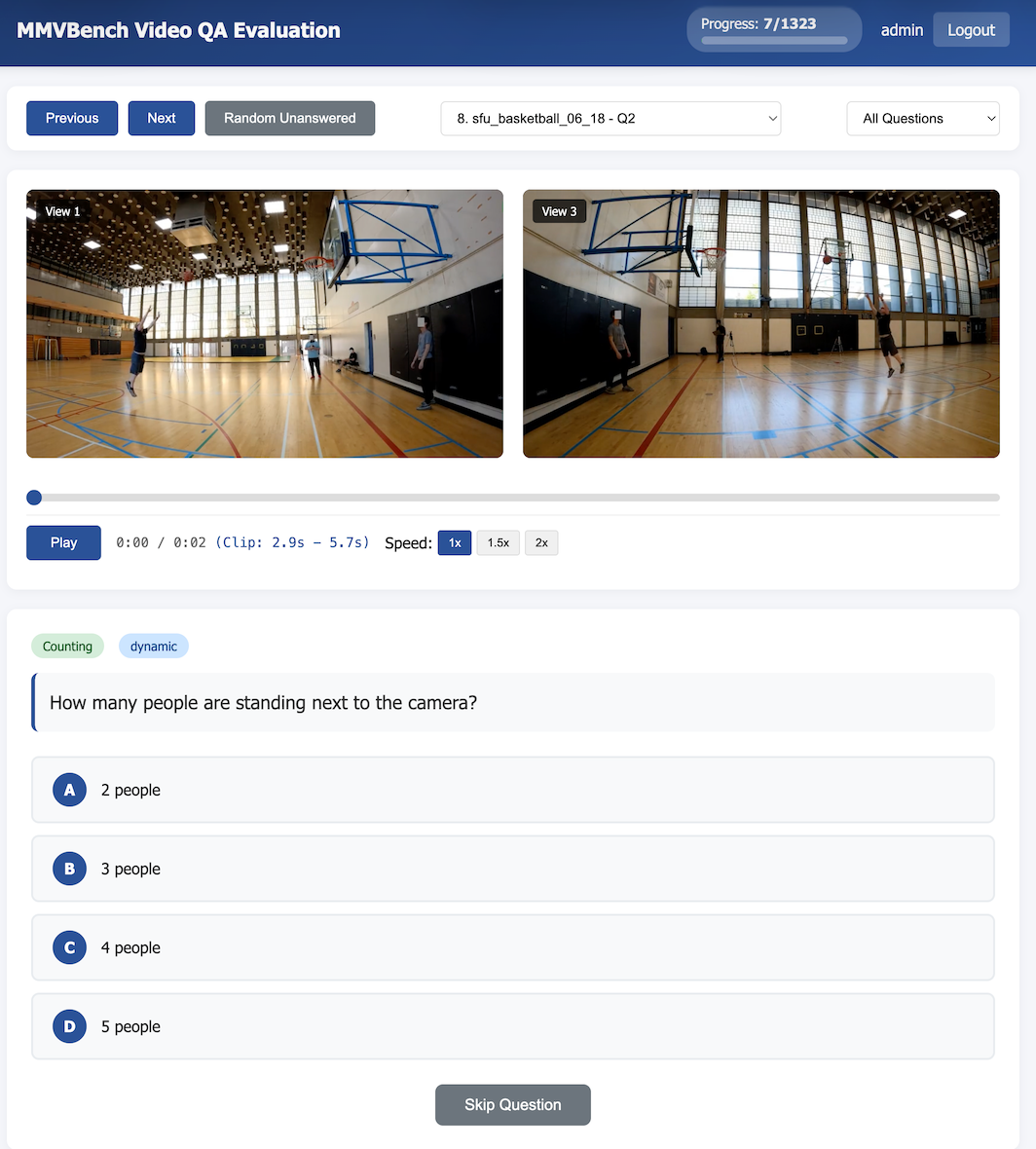}
    \caption{\textbf{Screenshot of the Human Evaluation Web GUI}.}
    \label{fig:human_evaluation_gui}
\end{figure}

\section{Benchmark Conversion Details}
\label{appendix:benchmark_conversion}

We convert each human-written open-ended annotation into a multiple-choice benchmark item while preserving the original reasoning target. Each annotation contains a selected set of views $\Vc$, an annotated temporal window $[t_s,t_e]$, a question $q$, a free-form answer $a$, and the original MVVBench semantic category $c$. The conversion stage outputs $(\Vc_{[t_s,t_e]}, q', \mathcal{O}, y, c)$, where $q'$ is the final benchmark question, $\Oc$ is the option set, and $y$ is the index of the correct option. Most items are converted into 4-way multiple-choice questions; binary items that cannot be faithfully rewritten remain 2-way. All model-based conversion steps operate on the same selected views and annotated temporal window used during labeling, rather than on the full raw video. Table~\ref{tab:prompt_benchmark_conversion} lists the exact prompts used in the prompt-based stages of this pipeline.

\paragraph{Answer-type classification.}
We first assign each QA pair to one of three answer types: \emph{binary}, \emph{counting}, or \emph{descriptive}. This step uses only the text pair $(q,a)$. Binary includes both yes/no questions and questions that ask the annotator to choose between exactly two alternatives; counting corresponds to integer-valued answers; and descriptive covers the remaining open-ended cases. This answer-type tag is used only to choose the downstream conversion rule and does not replace the original MVVBench category used in our benchmark analysis.

\paragraph{Binary-to-open conversion.}
Binary questions are retained only when a faithful open-ended reformulation is not available. Otherwise, we apply the prompted conversion in Table~\ref{tab:prompt_benchmark_conversion} to rewrite the original binary QA into an equivalent counting or descriptive QA using the original question, the original answer, and the associated clipped multi-view video. We accept a rewritten QA only if it preserves the intent of the original annotation, yields a concise non-binary answer, and remains specific enough to avoid degenerate outputs such as overly generic scene summaries or ambiguous counting targets. If no natural reformulation is available, we keep the original binary form and later convert it directly into a 2-option multiple-choice item. Using this procedure, 93 binary QAs
were rewritten into counting or descriptive QAs.

\paragraph{Counting questions.}
For counting items, we normalize the reference answer to its cardinal value and generate
distractors deterministically using the nearest integers around the ground-truth count.
Concretely, we collect the closest distinct non-negative integers to the correct count,
excluding the correct value itself, until we obtain three distractors. This produces
numerically plausible alternatives without requiring any additional video access. The final option text preserves the natural surface form of the answer (e.g., ``4 people'' rather than only ``4'').

\paragraph{Descriptive questions.}
Descriptive QAs require semantically plausible but incorrect distractors. Following the
generator--judge setup described in Sec.~3.2, we use one VLM to propose three candidate
distractors conditioned on the clipped multi-view video, the question, and the reference
answer, and a second VLM to evaluate the proposal. The generator is instructed to produce
alternatives that are mutually distinct, similar in specificity and surface form to the correct answer, and visually plausible in the scene, while still being unambiguously false. The judge rejects candidate sets that contain duplicate meanings, trivial negatives, multiple correct options, distractors unsupported by the video, or distractors that make the converted item noticeably easier than the original open-ended annotation. When a set is rejected, the generator is prompted again until an acceptable option set is obtained.

\paragraph{Final formatting and verification.}
After distractor generation, we combine the correct answer with the accepted distractors and randomly shuffle the option order. For binary questions that remain binary, yes/no questions become the option pair ``Yes''/``No'', while forced-choice questions preserve the two alternatives stated in the original annotation. Every converted item is then checked by a human verifier. The verifier confirms that (i) the labeled correct option is indeed correct, (ii) the remaining options are false but plausible, (iii) the converted item still reflects the intended cross-view reasoning challenge of the original annotation, and (iv) the wording is natural and unambiguous. Items that fail this check are edited or regenerated before entering the final benchmark release.

\begin{table}[!th]
\centering
\begin{minipage}{0.99\columnwidth}\vspace{0mm}    \centering
    \begin{tcolorbox} 
        \raggedright
        \small
\textcolor{GoogleBlue}{\textbf{Answer-type classification:}\\}
Classify the following question-answer pair into one of three categories.

Question: \{question\_text\}\\
Answer: \{answer\_text\}\\

Categories:\\
1. "binary" - The answer is Yes/No, OR the question asks to choose between exactly two options (e.g., "Which is closer, A or B?" with answer "A")\\
2. "counting" - The question asks "how many" or the answer is a number/count\\
3. "descriptive" - Open-ended questions that require a descriptive answer (not yes/no, not a simple count)\\

Respond with ONLY one word: "binary", "counting", or "descriptive" (nothing else).\\        
\textcolor{GoogleRed}{\textbf{Binary conversion:}\\}
You are helping convert binary QAs about videos into open-ended QAs.\\

Given a binary question and its answer (and the relevant video clip), rephrase it as an open-ended question that requires a descriptive or counting answer. You should rely on the original QA more than the video context, as it is a ground truth QA generated by a human annotator (expert). Try to maximally leverage the information in the original QA.\\

Note that the binary question may be yes/no question, or a question that you have to choose from two options that are not yes or no. Here is an example of a non-yes/no question: Q. "Which is the person in gray clothes closer to, the white cabinet or the gray box?" A. "White cabinet".\\

Original Question: \{question\_text\}\\
Original Answer: \{answer\}\\
Original Category: \{category\}\\

Guidelines:\\
1. If the question asks about a count (e.g., "Are there 3 people..."), convert to "How many..." with the actual number as the answer\\
2. The QA should not be ambiguous. For instance, the answer should not be "Multiple people" for the counting task. If it is, then you shoud rephrase the question to make it a descriptive question, so that the answer is no longer "Multiple people".\\
3. If the question asks about an action or state (e.g., "Is anyone moving..."), convert to "What is..." or "Who is..." with a descriptive answer \\
4. If the answer is "No", the new answer should reflect what is actually happening (the opposite or alternative). Use the video to determine what is actually happening.\\
5. Keep the context and specificity of the original question \\
6. The new answer should be concise but complete \\
7. Use the video context to provide accurate answers, when a good open-ended QA cannot be generated from the original QA alone. \\
8. Refrain from generating open-ended QAs such as "Describe the video" or "What is the main activity in the video?" Use questions that are more specific so that the answer can be concise and complete.\\

Examples:
- "Are there 2 people standing?" (Yes) → \{"new\_question": "How many people are standing?", "new\_answer": "2 people", "new\_category": "Counting"\}
- "Is anyone moving?" (No) → \{"new\_question": "What are the people doing?", "new\_answer": "Everyone is standing still", "new\_category": "Descriptive"\}
- "Are there 3 cameras?" (No, actually 4) → \{"new\_question": "How many cameras are there?", "new\_answer": "4 cameras", "new\_category": "Counting"\}\\

Output only the JSON object, nothing else.\\
\end{tcolorbox}
\caption{Prompts used for answer-type classification in \S\ref{subsec:benchmark_conversion}.}
\label{tab:prompt_benchmark_conversion}
\end{minipage}
\end{table}


\section{Human Performance Evaluation}
\label{appendix:human_perf_eval}

For human evaluation, we built a separate custom web interface that presented an expert annotator with clipped multi-view video segments together with the corresponding MVVBench question--answer pairs (Fig.~\ref{fig:human_evaluation_gui}). No time limit was imposed, allowing the annotator to pause and replay the videos as needed. Response times were recorded per question.

\paragraph{Overall statistics.}
On average, answering a single question took 60.34 seconds, which is generally longer than the response time of a typical VLM. This suggests that the reasoning required to answer MVVBench questions correctly is nontrivial, even for humans.

\paragraph{Per-category breakdown.}
Table~\ref{tab:human_detailed} reports human accuracy and average response time per category. Human performance is highest on relative distance with view index (90.12\%) and attribute identification (85.87\%), and lowest on relative pose (58.33\%). Notably, relative pose questions require by far the longest deliberation time (467.48s on average), roughly $10\times$ longer than attribute identification (34.68s), yet still yield the lowest accuracy. This suggests that relative camera pose reasoning is fundamentally difficult for humans as well, not merely a VLM-specific weakness.

\paragraph{Time--accuracy correlation.}
A striking pattern emerges when comparing response times for correct vs.\ incorrect answers: correctly answered questions take an average of 36.07 seconds, while incorrectly answered questions take 252.02 seconds---approximately $7\times$ longer. This indicates that difficult questions are not simply guessed quickly and incorrectly; rather, evaluators invest substantially more deliberation time on questions they ultimately answer wrong, suggesting these questions involve genuinely ambiguous or complex multi-view reasoning that resists resolution even with extended effort.

\paragraph{Breakdown by answer type.}
We also analyze performance by answer format. Binary questions are easiest (86.63\% accuracy, 34.76s average), followed by counting questions (77.13\%, 41.71s). Descriptive questions are the most challenging (73.70\%, 179.50s), reflecting the higher cognitive load of synthesizing free-form answers that require integrating evidence across views.

\begin{table}[h]
\caption{\textbf{Detailed human performance by category.} Cor: number correct, Tot: total, Acc: accuracy, Time: average response time in seconds.}
\centering
\small
\setlength{\tabcolsep}{5pt}
\renewcommand{\arraystretch}{1.15}
\begin{tabular}{@{} l r r r r @{}}
\toprule
Category & {Cor} & {Tot} & {Acc (\%)} & {Time (s)} \\
\midrule
Attribute identification & 480 & 559 & 85.87 & 34.68 \\
Counting & 276 & 358 & 77.09 & 40.92 \\
Attribute id.\ w/ view index & 90 & 121 & 74.38 & 44.79 \\
Relative pose & 70 & 120 & 58.33 & 467.48 \\
Relative distance & 73 & 84 & 86.90 & 58.63 \\
Relative distance w/ view index & 73 & 81 & 90.12 & 44.49 \\
\midrule
\textbf{Overall} & \textbf{1{,}062} & \textbf{1{,}323} & \textbf{80.27} & \textbf{60.34} \\
\bottomrule
\end{tabular}
\label{tab:human_detailed}
\end{table}

\section{Experimental Details}
\label{appendix:exp_details}

\begin{table}[!th]
\centering
\begin{minipage}{0.99\columnwidth}\vspace{0mm}    \centering
    \begin{tcolorbox} 
        \raggedright
        \small
\textcolor{GoogleBlue}{\textbf{Baseline:}\\}
You are given videos from multiple views of a scene. Answer the following multiple choice question by selecting the correct option.

Question: \{question\}

Options: \{options\}

Answer with: "Final Answer: X"\\        
\textcolor{GoogleRed}{\textbf{CoT:}\\}
You are given videos from multiple views of a scene. Answer the following multiple choice question.

Question: \{question\}

Options: \{options\}

Provide step-by-step reasoning, then end with: "Final Answer: X"\\
\textcolor{GoogleGreen}{\textbf{MV-Guided CoT:}\\}
You are given videos from multiple views of a scene. Answer the following multiple choice question.

Question: \{question\}

Options: \{options\}

Think step by step:
1. First, carefully observe what is happening in each view.
2. Identify the relevant information from the videos.
3. Reason through each option.
4. Select the best answer.

Important instructions:
- Each view shows the SAME scene from a DIFFERENT angle
- Pay attention to spatial relationships across views
- Consider what is visible in one view but not another
- The views are labeled (View 0, View 1, etc.) to help you reference them

CRITICAL: To answer correctly, you MUST cross-reference information across different views. What appears on the left in one view may appear on the right in another view due to camera positioning.

After your reasoning, provide your final answer as: "Final Answer: X"\\
\end{tcolorbox}
\caption{Prompts used for the experiments in \S\ref{sec:experiments}.}
\label{tab:prompt_cot}
\end{minipage}
\end{table}


\section{Details in TADSG}
\label{appendix:tadsg}

\paragraph{Design philosophy.}
Rather than asking the VLM to solve the full multi-view reasoning problem end-to-end, \textsc{TADSG} decomposes the problem into stages: \emph{establish cross-view identity} using specialized vision models, then \emph{compress the relevant evidence} into a structured intermediate representation, and finally \emph{answer from the graph} rather than from raw video. The approach is intentionally conservative: it activates only when cross-view correspondence is reliable, and falls back to the best prompting baseline otherwise. This design is motivated by the observation that forcing a low-confidence graph degrades performance relative to the baseline.

\paragraph{Cross-view identity establishment.}
Given a multi-view clip, we first run a person detector (YOLOv8~\cite{yolov8_ultralytics}) on uniformly sampled frames across all views and select an \emph{anchor frame}---the temporal moment at which the most people are jointly visible across the most views. At the anchor frame, we crop each detected person, extract appearance embeddings with a person re-identification model (OSNet~\cite{zhou2019osnet}), and greedily cluster detections \emph{across views} based on cosine similarity. Each resulting cluster defines a global identity label (e.g., $P_1$, $P_2$), establishing explicit cross-view correspondence that the VLM cannot reliably infer on its own.

\paragraph{Confidence gating.}
A key design choice is that \textsc{TADSG} does not activate unconditionally. Two sequential gates must be passed before the dynamic scene graph is constructed:
\begin{enumerate}
    \item \textbf{Anchor coverage gate.} The fraction of detected persons successfully matched across views must exceed a category-dependent threshold. If anchor correspondence is too sparse, the method falls back to the prompting baseline.
    \item \textbf{Target similarity gate.} After the VLM resolves the question's natural-language person reference (e.g., ``the person in black'') to a global identity, the minimum pairwise ReID similarity within that identity's cluster must exceed a threshold. If the target's cross-view evidence is unreliable, the method again falls back.
\end{enumerate}
This selective activation ensures that the full-benchmark gain comes from a smaller number of high-confidence activations rather than from uniformly applying a potentially noisy graph.

\paragraph{Dynamic scene graph construction.}
For questions that pass both gates, we construct a \emph{query-conditioned, target-centric} dynamic scene graph (DSG). Unlike conventional scene graphs that describe the entire scene~\cite{ji2020action}, our DSG is built exclusively around the resolved target entity and the relations relevant to the question. The graph follows a fixed five-block schema:
\begin{itemize}
    \item \textbf{Target Node}: the identity and role of the queried person.
    \item \textbf{Anchor Observations}: per-view appearance, position, and size at the anchor moment.
    \item \textbf{Temporal Observations}: per-view motion, action, and visibility over time.
    \item \textbf{Cross-View Relations}: identity continuity, relative camera distance, and direction mapping across views.
    \item \textbf{Question-Relevant Facts}: compressed facts that the final answer stage should use.
\end{itemize}
The VLM generates the graph from a target-focused evidence pack consisting of cropped identity montages, target-highlighted anchor frames, and the full video clips, guided by the global identity labels and the correspondence hint from the ReID stage.

\paragraph{Graph-grounded answering.}
In the final stage, the VLM answers the question using the dynamic scene graph as its \emph{primary evidence}. The target-focused anchor images serve only to resolve minor remaining ambiguities. By externalizing cross-view identity, spatial relations, and temporal dynamics into a structured text before the answer is produced, \textsc{TADSG} reduces the reasoning problem from ``infer everything from raw pixels'' to ``reason over a compact, pre-verified summary.'' This decomposition is reminiscent of how scene graphs have been used as intermediate representations for spatial reasoning in embodied QA~\cite{saxena2024grapheqa} and video understanding~\cite{ji2020action}, but adapted here for the multi-view setting with explicit tool-augmented identity grounding.

\section{RLVR post-training}
\label{appendix:rlvr}

For each training example, we used the two synchronized camera views associated with the question and trimmed each video to the annotated temporal interval. From each trimmed clip, we sampled four frames per view and converted them into a labeled multi-view temporal panel with two rows (one per view) and four columns (temporal order). This representation preserved both cross-view correspondence and short-term temporal information while keeping RL training computationally manageable.

Training used \texttt{trl} for DAPO-based RL, LoRA adapter tuning, and \texttt{bitsandbytes} for 4-bit NF4 model loading. We trained LoRA adapters with rank 16, LoRA alpha 32, and dropout 0.05 on the main attention and MLP projection layers.
We set learning rate to be \(2\times 10^{-6}\), ran 1200 update steps, per-device batch size 1, gradient accumulation 1, 4 generations per prompt, maximum completion length
48, temperature 0.9, top-\(p\) 0.95, top-\(k\) 40, repetition penalty 1.02, and KL coefficient \(\beta=10^{-3}\). We used group reward scaling, and 2 GRPO inner iterations.



\end{document}